\documentclass{article}

\usepackage[preprint]{neurips_2024}

\usepackage[utf8]{inputenc} 
\usepackage[T1]{fontenc}    
\usepackage{hyperref}       
\usepackage{url}            
\usepackage{booktabs}       
\usepackage{amsfonts}       
\usepackage{nicefrac}       
\usepackage{microtype}      
\usepackage{xcolor}         

\usepackage{graphicx}

\usepackage{enumitem}

\usepackage{amssymb}

\usepackage{multirow}

\usepackage{subfig}

\def\summcorpus{MILDSum}

\newcommand{\new}[1]{\textcolor{black}{#1}}
\newcommand{\addnew}[1]{\textcolor{black}{#1}}

\title{Advantages of Domain Knowledge Injection for Legal Document Summarization: A Case Study on Summarizing Indian Court Judgments \\ in English and Hindi}

\author{%
  Debtanu Datta \\
  Department of Mathematics \\
  Indian Institute of Technology Kharagpur \\
  Kharagpur, India 721302 \\
  \texttt{debtanudatta04@gmail.com} \\
  \And
  Rajdeep Mukherjee \\
  Department of Computer Science and Engineering \\
  Indian Institute of Technology Kharagpur \\
  Kharagpur, India 721302 \\
  \texttt{rajdeep1989.iitkgp@gmail.com} \\
  \AND
  Adrijit Goswami \\
  Department of Mathematics \\
  Indian Institute of Technology Kharagpur \\
  Kharagpur, India 721302 \\
  \texttt{goswami@maths.iitkgp.ac.in} \\
  \And
  Saptarshi Ghosh \\
  Department of Computer Science and Engineering \\
  Indian Institute of Technology Kharagpur \\
  Kharagpur, India 721302 \\
  \texttt{saptarshi@cse.iitkgp.ac.in} \\
}

\begin{document}

\maketitle

\begin{abstract}
Summarizing Indian legal court judgments is a complex task not only due to the intricate language and unstructured nature of the legal texts, but also since a large section of the Indian population does not understand the complex English in which legal text is written, thus requiring summaries in Indian languages. In this study, we aim to improve the summarization of Indian legal text to generate summaries in both English and Hindi (the most widely spoken Indian language), by injecting domain knowledge into diverse summarization models. 
We propose a framework to enhance extractive neural summarization models by incorporating domain-specific pre-trained encoders tailored for legal texts. Further, we explore the injection of legal domain knowledge into generative models (including Large Language Models) through continual pre-training on large legal corpora in English and Hindi. Our proposed approaches achieve statistically significant improvements in both English-to-English and English-to-Hindi Indian legal document summarization, as measured by standard evaluation metrics, factual consistency metrics, and legal domain-specific metrics. Furthermore, these improvements are validated through domain experts, demonstrating the effectiveness of our approaches. 
\end{abstract}


\section{Introduction} 
\label{sec:intro}

Legal court judgment summarization is a challenging task due to the intricate complexity of legal texts which are often unorganized and lengthy~\citep{Deroy-llm-legal-summarization,Dan2023EnhancingLJ,Nguyen2023KeywordbasedAM,Moro2023MultilanguageTL}. 
There are additional complexities in Indian legal text summarization -- most legal texts in India are written in complex English due to historical reasons, but a large section (more than 80\%) of the Indian population lacks command of English. This creates a significant language barrier, especially when individuals from financially weaker sections confront legal situations. 
Hence, summarizing legal documents in Indian languages is essential to give access to justice to a broader population. 

Despite the significance of the problem, to our knowledge, there is only one summarization dataset that provides Indian legal documents along with their summaries in both English and an Indian language (Hindi, the most widely spoken Indian language) -- \summcorpus{} that was introduced in our prior work~\citep{datta-etal-2023-mildsum}.
However, there is a lot of scope for improvement in the performance of existing summarization models on this dataset (as reported in~\citet{datta-etal-2023-mildsum}).
In particular, there has not been much attempts to improve legal document summarization by injecting domain knowledge into summarization models. 
For instance, while there exist large Indian legal corpora in English and Hindi, along with BERT-based pre-trained models tailored for Indian legal texts~\citep{InLegalBERT_paper}, these resources have \textit{not} yet been leveraged to enhance the performance of legal document summarization in English and Hindi. 

\new{
In this study, we address this gap by proposing novel approaches that inject legal domain-specific knowledge into diverse summarization models, towards improving legal document summarization. Our key contributions are as follows: }

\begin{enumerate}[leftmargin=*]

    \item \new{\textbf{Domain knowledge injection in Extractive Summarization models:} We introduce a novel framework for integrating domain knowledge into an extractive encoder-only summarizer by incorporating domain-specific pre-trained encoders.
    }

    \item \new{\textbf{Domain Knowledge Injection in Generative Models:} We enhance both monolingual (English-to-English, EN-to-EN) and cross-lingual (English-to-Hindi, EN-to-HI) summarization of Indian legal documents by injecting legal domain knowledge into generative models through continual pre-training on large legal corpora.
    }

    \item \new{\textbf{Exploration of Pre-Training Strategies:} We explore both resource-intensive and resource-efficient, i.e., with and without memory-efficient GaLore~\citep{galore_paper} training strategies to assess the effectiveness in our setup, along with the impact of pre-training with varying corpus sizes. Our experiments demonstrate that resource-efficient techniques can achieve nearly equivalent results in legal document summarization.
    }

    \item \new{\textbf{Investigation of Cross-Lingual Transfer Effects by Comprehensive Multilingual Pre-Training experiments:} We perform comprehensive multilingual pre-training experiments over English-only, Hindi-only, and an equal mix of English and Hindi corpora to analyze the impact of the cross-lingual transfer effects in the downstream English-to-Hindi summarization task.
    }

    \item \new{\textbf{Comprehensive Model Comparisons:} We conduct comprehensive evaluation of our approaches across diverse model architectures, including Encoder-only, Decoder-only, and Encoder-Decoder models, with almost comparable model sizes, highlighting the effectiveness of domain knowledge injection across different types of models (both summarization-specific models as well as general-purpose LLMs).
    }

\end{enumerate}

Our approaches demonstrate substantial (statistically significant) performance gains over the state-of-the-art (SOTA). 
\new{Specifically, we outperform the SOTA baselines on \summcorpus{} benchmark for both English-to-English and English-to-Hindi summarization, achieving 20--23\% improvement in ROUGE-2 F1 and 15--19\% improvement in ROUGE-L F1 scores. 
We also demonstrate that domain knowledge injection helps in making the summaries more factually consistent.
We validate the superior quality of summaries generated by our models through qualitative assessment by a Law domain expert, further confirming the effectiveness of our approaches in generating high-quality legal summaries.
Additionally, we compare our models against \textit{GPT-4}, demonstrating that \textit{GPT-4} still struggles with domain-specific and cross-lingual tasks.
}
These results highlight the effectiveness of our domain knowledge injection approaches and their potential to enhance legal text summarization. It can be noted that, though this work focuses on summarization in English and Hindi, the approaches developed in this work are applicable to enhance legal text summarization in other languages as well. 

\new{
The rest of the paper is organized as follows. Section~\ref{sec:related} discusses prior work about legal document summarization. Sections~\ref{subsec:pt-data-details} and \ref{subsec:summ-data} describe the legal corpora utilized for pre-training and fine-tuning experiments, respectively. The evaluation metrics are detailed in Section~\ref{subsec:EvalMetrics}.
Section~\ref{sec:expResult} presents our experiments and results across different model architectures.
Specifically, Section~\ref{subsec:supExtSum} describes the extractive Encoder-only models, Section~\ref{subsec:abs-encodeco-summ} explores generative Encoder-Decoder models for both EN-to-EN and EN-to-HI summarization, and Section~\ref{subsec:abs-deco-summ} presents the experiments and results of generative Decoder-only model. 
The hyperparameter configurations for all experiments are detailed in Section~\ref{subsec:hyper-param}, followed by Section~\ref{subsec:gpt-4}, which compares the performances of \textit{GPT-4} against our models.
Section~\ref{sec:expert-eval} focuses on detailed expert evaluation.  
Finally, Section~\ref{sec:conclu} concludes the paper, summarizing the key findings of this study.
}

\section{Related Works} 
\label{sec:related}


\subsection{Strategies for domain knowledge injection}

Researchers are increasingly focusing on strategies to incorporate domain-specific knowledge in Large Language Models (LLMs) to enhance their performance in domain-specific tasks. For instance, \citet{moiseev-etal-2022-skill} infused structured knowledge into encoder-decoder models by training them directly on factual triples of knowledge graphs. \citet{xu-etal-2023-kilm} introduced a generative infilling objective to enhance models by integrating entity-related knowledge through continued pre-training. Also, \citet{k_adapter_paper} proposed a neural adapter framework, allowing knowledge infusion without altering the pre-trained model's original parameters, thus enabling continuous updates for different types of knowledge via neural adapters.

However, to our knowledge, no prior work has attempted domain knowledge injection to enhance the summarization of long legal documents. 

\subsection{Improving legal document summarization}

Given the intricate nature of legal texts and the scarcity of legal datasets, improving legal document summarization is a challenging task. 
To this end, \citet{Moro2023MultilanguageTL} address the scarcity of labeled datasets by applying a transfer learning approach, combining extractive and abstractive techniques to enhance summarization performance in low-resource legal cases. 
Also, \citet{Dan2023EnhancingLJ} propose an integrated approach that leverages both semantic and structural features to improve the quality of legal judgment summaries.
\citet{Nguyen2023KeywordbasedAM} introduce a keyword-based augmentation method to guide abstractive summarization models by infusing keywords that help capture essential information from lengthy legal documents.

In the context of the Indian legal domain, \citet{Jain2023ASI} develop DCESumm, a hybrid sentence scoring approach combining supervised sentence-level summary relevance prediction with unsupervised clustering-based document-level score enhancement, to give a better extractive summary of Indian legal documents. 
\citet{Bhattacharya2021IncorporatingDK} introduce DELSumm, an unsupervised algorithm that systematically integrates expert-guided legal domain knowledge into an optimization-based framework, outperforming several supervised models. \citet{Deroy2023EnsembleMF} demonstrate that intelligently ensembling multiple base summarizers can produce superior summaries for legal judgments compared to individual models.

However, all these prior works have considered summaries in English only.
To the best of our knowledge, only our prior work~\citep{datta-etal-2023-mildsum} recently introduced a multilingual Indian legal summarization dataset, \summcorpus{}, that contains Indian court judgments in English, along with summaries in both English and Hindi (details are discussed in Section~\ref{subsec:summ-data}). We benchmarked several summarization models over the dataset. 
\new{The state-of-the-art (SOTA) on the \summcorpus{} dataset, as reported in~\citep{datta-etal-2023-mildsum}, were \textit{SummaRuNNer}~\citep{summarunner} for EN-to-EN summarization and \textit{CrossSum-mT5}~\citep{bhattacharjee2023crosssum} for EN-to-HI summarization. 
we outperform both of these models through domain knowledge injection in this paper.
}


\vspace{2mm}
\noindent \textbf{Present work as an extension of our prior work:} 
\new{As stated earlier, the \summcorpus{} dataset used in this paper, was introduced in our prior work~\citep{datta-etal-2023-mildsum}, where we primarily focused on dataset creation and benchmarking existing summarization models. 
Our present study contains substantial enhancements over the prior work, which are as follows: 
(1)~In the present work, we enhance both monolingual (EN-to-EN) and cross-lingual (EN-to-HI) summarization by injecting legal domain knowledge into generative models via continual pre-training on large legal corpora. 
(2)~We made significant improvement over the SOTA baseline for EN-to-EN legal summarization on \summcorpus{} (reported in our prior work~\citep{datta-etal-2023-mildsum}) by architectural modification through domain-specific pre-trained Encoders. 
(3)~We explore resource-intensive and resource-efficient pre-training strategies, along with comprehensive evaluation across diverse model architectures.
(4)~Further, we investigate cross-lingual transfer effects via deep multilingual pre-training analysis. 
(5)~We now introduce a new legal domain-specific evaluation metric and added factual consistency metrics for better assessment of legal judgment summaries. These metrics were not explored in our prior work. 
(6)~The present work also includes a human evaluation of our summaries, which further validates our approaches.
Overall, these enhancements extend our prior work from dataset creation to a more methodologically rigorous study, providing deeper insights for improving legal document summarization in various languages.
}
\section{Datasets and Evaluation metrics}
\label{sec:datasets}

This section describes the datasets used for continual pre-training (for domain knowledge injection) and summarization. The evaluation metrics used for assessing the model-generated summaries are also discussed in this section.

\subsection{Corpora for Pre-training}
\label{subsec:pt-data-details}

We utilize the following two existing Indian legal corpora to pre-train the models in English and Hindi for the purpose of domain knowledge injection.

\noindent \textbf{(1)~InLegalBERT-PT:} For English, we utilize small subsets (around 27K and 68K documents) from the pre-training corpus of the \textit{InLegalBERT} model~\citep{InLegalBERT_paper}, which is a comprehensive collection of court case judgments from the Supreme Court and various High Courts of India, along with Indian Government statutes.
The judgments span diverse legal areas, including civil, criminal, and constitutional law, making it a rich resource for injecting domain-specific legal knowledge. 
For more details regarding this pre-training corpus, please see the paper by~\citet{InLegalBERT_paper}.

\noindent \textbf{(2)~Bail Corpus:} For Hindi, we utilize small subsets (around 17K and 34K documents) from the corpus of~\citet{kapoor-etal-2022-hldc} that contains 
around 177K court judgments in Hindi from bail application cases decided in district courts in the state of Uttar Pradesh, India. 
For further details about this Bail corpus, see~\citet{kapoor-etal-2022-hldc}.


\addnew{We experiment with different pre-training corpus sizes to systematically analyze how pre-training with varying corpus sizes influence model performance on the downstream summarization task.}
Our decision to select a small subset from both the above corpora aligns with our quest to understand whether it is possible to gain performance improvements in legal data summarization by means of injecting domain knowledge with \textit{limited but relevant} domain-specific but task-agnostic data.
As reported in later sections (e.g., Table~\ref{tab:pt-corpus-results}), our experimental observations support our decision.




\subsection{Dataset for Summarization Experiments}
\label{subsec:summ-data}

We choose the \textbf{\summcorpus{}} dataset (introduced in our prior work~\citep{datta-etal-2023-mildsum}) for summarization experiments,
since it is currently the only dataset that provides summaries of Indian court judgments in both English and Hindi, which enables us to check the effects of pre-training in multiple languages.
The dataset is specifically curated for the summarization task within the Indian legal domain and comprises 3,122 Indian court judgments in English, along with their summaries in both English and Hindi. These summaries are drafted by Law practitioners which ensure domain relevance and accuracy. 


\vspace{2mm}
\noindent \textbf{Dataset Statistics:}
The dataset has three splits -- train, validation, and test. 
\addnew{Statistics of this dataset is given in Table~\ref{tab:Dataset-stats}.}
In our experimental setup, the train and validation splits are employed to fine-tune the models, while the test split is reserved exclusively for evaluating the models' inference performance over the downstream summarization task.

\begin{table}[t]
\caption{Statistics of the \summcorpus{} dataset.}
\label{tab:Dataset-stats}
\centering
\begin{tabular}{lccccc}
    \toprule
    \multirow{2}{*} {\textbf{Split}} & \multirow{2}{*} {\textbf{\#datapoints}} & \multicolumn{3}{c}{\textbf{Avg. \#words}} \\ \cmidrule{3-5}
    & & \textbf{Doc} & \textbf{EN\_Sum} & \textbf{HI\_Sum} \\ 
    \midrule
    train & 2,185 & 4655 & 753 & 674 \\
    validation & 469 & 5319 & 758 & 687 \\
    test & 468 & 4473 & 760 & 664 \\
    \bottomrule
    \end{tabular} %
\end{table}


\vspace{2mm}
\noindent \textbf{Extractiveness of the English summaries:} One notable characteristic of \summcorpus{} is the extractive-ness of the English summaries -- the summaries frequently contain sentences directly quoted from the judgments. 
Hence, the dataset shows a high \textit{Extractive Fragment Coverage} of 0.90 and a high \textit{Extractive Fragment Density} of 24.42~\citep{datta-etal-2023-mildsum}. The \textit{Coverage} measures the proportion of words in the summary that are part of extractive fragments from the judgment, while \textit{Density} captures the average length of these extractive fragments~\citep{grusky-etal-2018-newsroom}. 
For more details about the \summcorpus{} dataset, we refer the readers to our prior paper~\citep{datta-etal-2023-mildsum}.

\subsection{Evaluation Metrics}
\label{subsec:EvalMetrics}

\new{In the subsequent sections, we evaluate the quality of model-generated summaries across two aspects -- 
(i)~\textbf{Standard relevance metrics:} measures the \textit{matching} with the gold-standard / reference summaries contained in the dataset, and 
(ii)~\textbf{Factual consistency metrics}: measures the \textit{consistency} of the abstractive summaries with the input source documents. 
To this end, we employ the following evaluation metrics. 
}

\subsubsection{Metrics for matching with reference summaries}

To measure the match of a generated summary for a document with
the gold standard / reference summary of the same document, the following metrics have been used.

\begin{itemize}[leftmargin=*]

\item \textbf{ROUGE} (Recall-Oriented Understudy for Gisting Evaluation)~\citep{lin-2004-rouge} evaluates the overlap between the model-generated and reference summaries. It is used for syntactic evaluation of generated summaries. Specifically, ROUGE-2 (R-2) captures the overlap of bi-grams, while ROUGE-L (R-L) measures the longest matching sequence of words using the Longest Common Subsequence (LCS). For multilingual evaluation (covering both English and Hindi), we utilized the \textit{multilingual\_rouge\_scoring} library\footnote{\url{https://github.com/csebuetnlp/xl-sum}}.
Specifically, we report F1 scores of ROUGE-2 (R-2) and ROUGE-L (R-L) in our experiments.

\item \textbf{InLegal-BERTScore} and \textbf{BERTScore} leverage embeddings from BERT to compute the similarity between the token-level embeddings of the generated and reference summaries. Unlike the ROUGE metric, which focuses on exact matches or n-gram overlaps, BERTScore evaluates the semantic similarity between the reference summary and the model-generated summary~\citep{Zhang2020BERTScore}. It aligns more closely with human judgments. We used the official implementation\footnote{\url{https://github.com/Tiiiger/bert_score}} of BERTScore calculation.
For English summaries, we introduce a novel \textit{legal domain-specific BERTScore metric}, referred to as InLegal-BERTScore (abbreviated as InL-B-S), which leverages \textit{InLegalBERT}~\citep{InLegalBERT_paper} -- a BERT model pre-trained on a large Indian legal corpus -- to capture legal nuances in English summaries. 
For Hindi summaries, since there does not exist any legal domain-specific BERT model supporting Hindi, we use \textit{multilingual-BERT}\footnote{\url{https://huggingface.co/google-bert/bert-base-multilingual-cased}} for the \textit{BERTScore} (abbreviated as B-S) calculation of Hindi summaries. Specifically, we report the F1 scores of InL-B-S for English and F1 scores of B-S for Hindi.

\end{itemize}



\subsubsection{Metrics for factual consistency of generated summaries}

It is important to evaluate the consistency of legal summaries, since generative models may hallucinate while summarizing~\citep{Deroy-llm-legal-summarization}. To assess the factual consistency of English summaries, we consider the following metrics. 

\begin{itemize}[leftmargin=*]

\item \textbf{SummaCONV}~\citep{summac_paper}, which uses Natural Language Inferencing (NLI) to score each sentence in the summary, indicating the likelihood that the summary sentence logically follows from the sentences in the source document~\citep{summac_paper}. 
In NLI, one sentence is treated as the \textit{hypothesis} and another as the \textit{premise}, with the model assigning a score indicating how likely the \textit{hypothesis} logically follows from the \textit{premise}. For a given (document, summary) pair, SummaCONV (abbreviated as SummaC) computes NLI scores for each sentence in the model-generated summary, indicating the likelihood that the summary sentence logically follows from the sentences in the source document. The individual NLI scores are aggregated to produce a single SummaC score for the (document, summary) pair, with higher scores indicating better consistency between the summary and the document. We used the official implementation\footnote{\url{https://github.com/tingofurro/summac}} of SummaC to obtain the scores for all model-generated summaries.

\item \textbf{NEPrec} (abbreviated as N-P): Named entities, such as names of individuals, organizations, and locations, are critical in legal case judgments, as any misrepresentation of such entities can lead to significant information loss or factual inaccuracies. 
We check the consistency of named entities using NEPrec, which calculates the fraction of named entities in the summary that are also present in the original document. 
Named entities are identified in English text using the Spacy toolkit.\footnote{\url{https://spacy.io/}}
While the N-P metric provides valuable insights into entity consistency, its accuracy depends on the effectiveness of the named entity recognition model.

\end{itemize}


\new{We report the values of all metrics in the range [0, 100], where a higher value represents better summary in terms of matching with the reference summary and consistency with the source document. 
All metrics are averaged over all documents in the test-split of the dataset.}

\new{Apart from these automatic evaluation metrics, we also performed a human evaluation through a Law expert, as detailed later in the paper (Section~\ref{sec:expert-eval}).}



\section{Experiments and Results}
\label{sec:expResult}

In this section, we describe how we incorporate legal domain knowledge into different types of models -- extractive Encoder-only, generative Encoder-Decoder models, and Decoder-only models -- to enhance their summarization capabilities in the Indian legal domain for both English and Hindi languages.
\new{We also compare our models against \textit{GPT-4} on both monolingual (English-to-English) and cross-lingual (English-to-Hindi) summarization tasks.}


\begin{figure}[t]
\centering
    \subfloat[\textit{SummaRuNNer}]{
            \includegraphics[width=0.48\textwidth, height = 5.5cm]{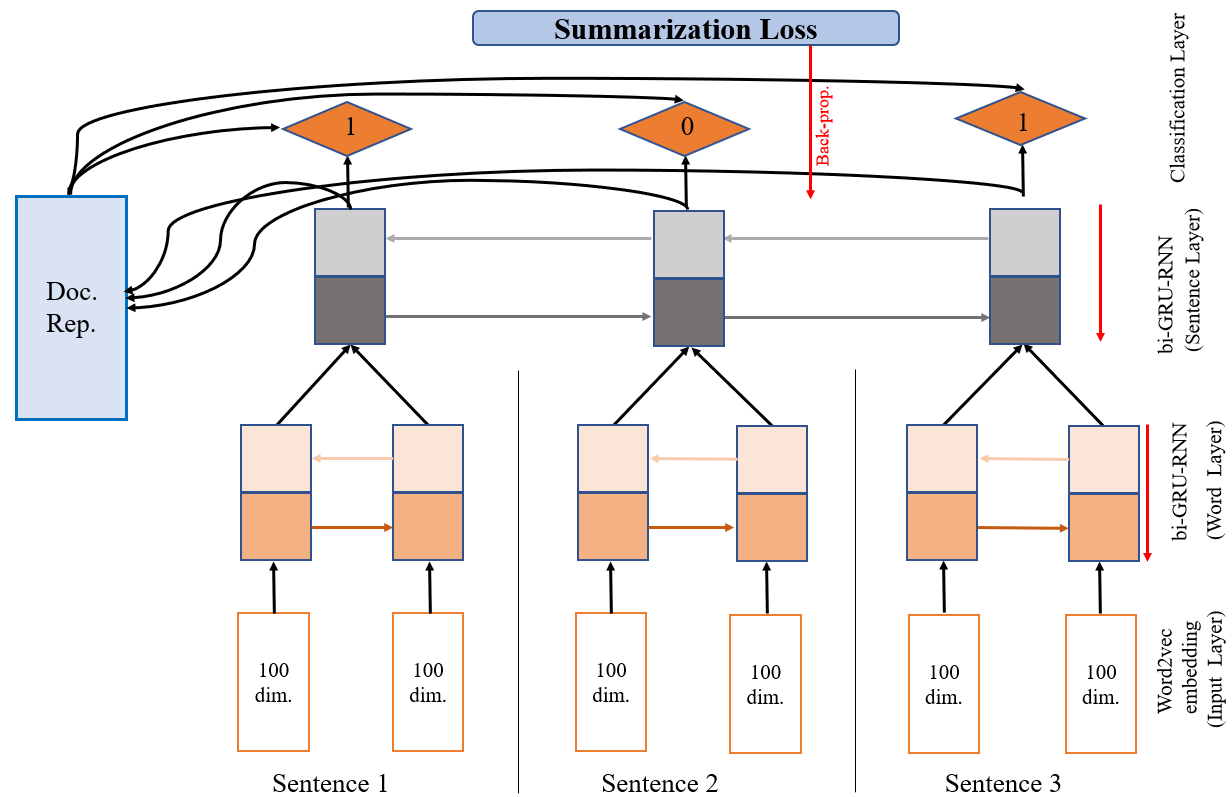}
    }
    \subfloat[\textit{InLegalSumExt}]{
            \includegraphics[width=0.48\textwidth, height = 5.5cm]{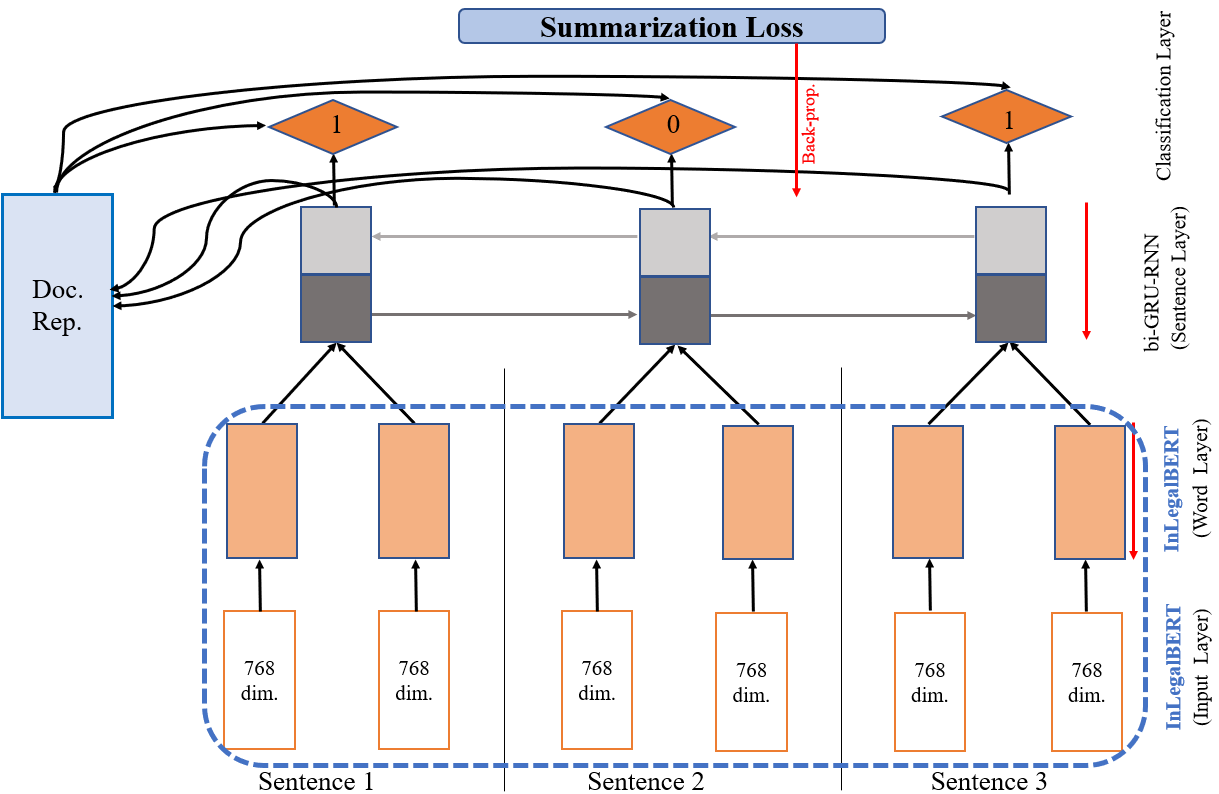}
    }
    \caption{Architectures of \textit{SummaRuNNer} and \textit{InLegalSumExt}}
\label{fig:archi_fig}
\end{figure}

\subsection{Extractive Encoder-only Models}
\label{subsec:supExtSum}

Here, our objective is to inject legal domain knowledge into a supervised extractive summarizer to enhance its domain-specific summarization performance. 
To this end, we select \textit{SummaRuNNer}~\citep{summarunner}, which delivered the best performance on the \summcorpus{} dataset in our prior work~\citep{datta-etal-2023-mildsum}. 
\textit{SummaRuNNer} frames summarization as a binary classification problem, where each sentence is classified as either part of the summary or not.

\vspace{2mm}
\noindent \textbf{Injecting domain knowledge into \textit{SummaRuNNer}:} 
We propose a generic approach to inject domain knowledge into \textit{SummaRuNNer} by modifying its two-layer bi-directional GRU structure.
In the original model, words are initialized with 100-dim word2vec embeddings trained on the CNN/Daily Mail corpus, and the first layer is a bi-GRU structure that generates contextualized word representations that are then average-pooled to obtain sentence embeddings~\citep{summarunner}. 
We replace the first layer with \textit{InLegalBERT}\footnote{\url{https://huggingface.co/law-ai/InLegalBERT}}~\citep{InLegalBERT_paper}, a BERT-based model pre-trained on Indian legal text.
Words are now initialized with 768-dim \textit{InLegalBERT} embeddings that provide domain-specific representations, which are then passed through the second bi-GRU layer. 
This architectural enhancement ensures that \textit{SummaRuNNer} leverages legal domain knowledge effectively. We refer to this modified model as \textit{InLegalSumExt}.
\new{Figure~\ref{fig:archi_fig} shows the architectural diagrams of the original \textit{SummaRuNNer} and \textit{InLegalSumExt} models, 
illustrating the flow of data through both architectures, and highlighting the modifications that enhance the summarization performance of \textit{InLegalSumExt}.
}

\begin{figure}[t]
\centering
\includegraphics[width=0.98\textwidth]{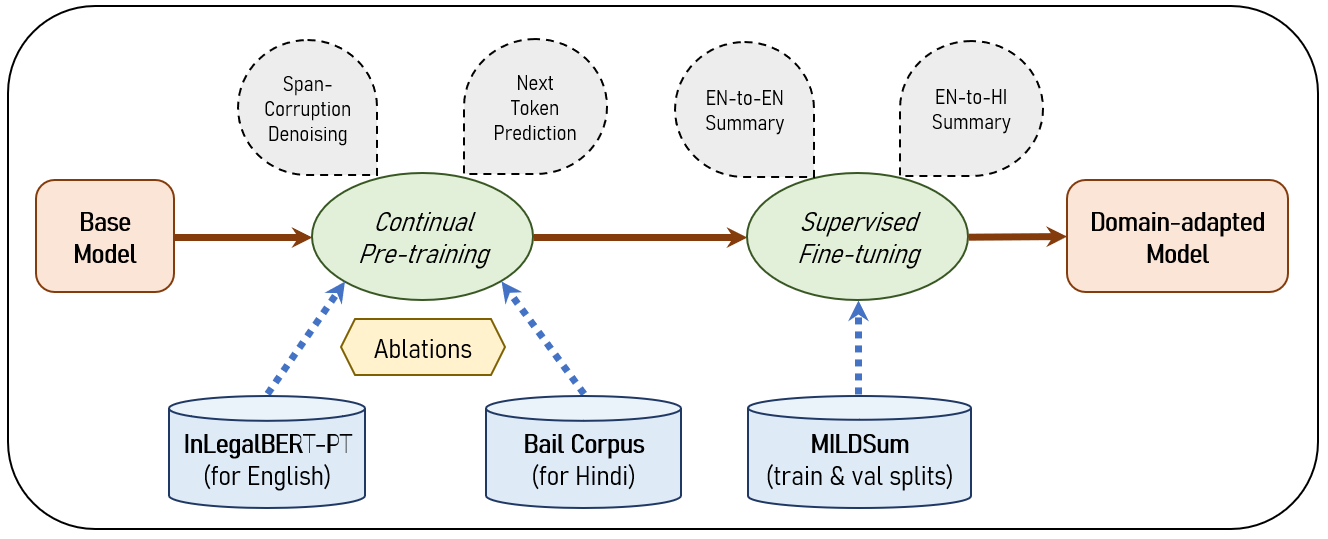}
\caption{We conduct continual pre-training over subsets of InLegalBERT-PT (for English) and Bail Corpora (for Hindi), then supervised fine-tuning over train and validation splits of \summcorpus{}.}
\label{fig:diagram_cpt_sft}
\end{figure}

\vspace{2mm}
\noindent \textbf{Training Supervised Extractive Models:} Supervised extractive methods like \textit{SummaRuNNer} require purely extractive reference summaries for training (for the binary classification task of selecting sentences for the summary). For training, every sentence in the document must be labeled as 1 (suitable for inclusion in the extractive summary) or as 0 (not suitable).
Since the reference summaries in \summcorpus{} are not purely extractive, it is necessary to convert them into extractive form. 
To this end, we follow the same method as in~\citep{datta-etal-2023-mildsum} to create the purely extractive reference summaries from the train split of \summcorpus{}. 
\addnew{Briefly, it is an unsupervised approach where we identify/select those sentences from a document that maximize the ROUGE score with respect to the reference (gold standard) abstractive summary.
In this regard, for a (document, summary) pair, label~1 is assigned to those document sentences that greedily (taken one at a time incrementally) maximize the ROUGE-2 overlap with the gold-standard reference summary. 
We stop until none of the remaining document sentences improves the ROUGE-2 score upon addition to the current extractive summary.
Finally, the rest of the sentences in the document are assigned label 0.
}
Both \textit{SummaRuNNer} and \textit{InLegalSumExt} models are trained using this approach.


\vspace{2mm}
\noindent \textbf{Summarization results:} Table~\ref{tab:EngSummResults} compares the performance of the vanilla \textit{SummaRuNNer} and our proposed \textit{InLegalSumExt} on the test split of \summcorpus{}.
Substantial performance improvements are observed for \textit{InLegalSumExt} --  
approx. 20\% improvement in ROUGE-2 F1 and 15\% improvement in ROUGE-L F1 scores compared to the best results reported by~\citet{datta-etal-2023-mildsum} for English summaries. 
In fact, \textit{InLegalSumExt} performs \textit{statistically significantly better} than \textit{SummaRuNNer} according to all the metrics of ROUGE-2 F1, ROUGE-L F1 and BertScore, according to both \textit{Wilcoxon Signed-Rank} and \textit{Mann-Whitney-U} tests at 99\% confidence interval.
This highlights the effectiveness of our architectural modifications and domain knowledge integration.

It can be noted that extractive summarization models outperform abstractive models in Table~\ref{tab:EngSummResults} since the summaries in \summcorpus{} contain a large number of long extractive fragments (quotes) directly taken from the judgment documents (as also reported in~\citep{datta-etal-2023-mildsum}). This extractive nature of \summcorpus{} was detailed in Section~\ref{subsec:summ-data}.

\begin{figure}[t]
\centering
    \includegraphics[width=\linewidth, height = 3cm]{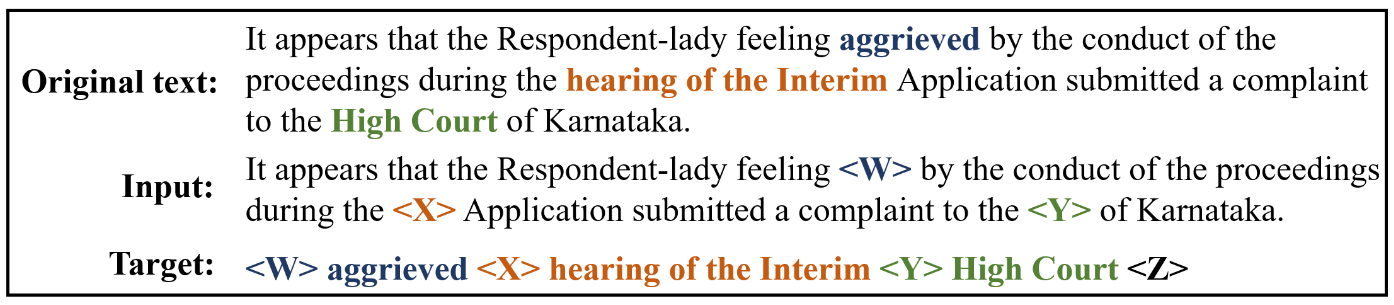}
    \caption{Example of sample data in English for span-corruption denoising pre-training objective.}
    \label{fig:pt-en-example}
    \vspace{2mm}
    \includegraphics[width=\linewidth, height = 3cm]{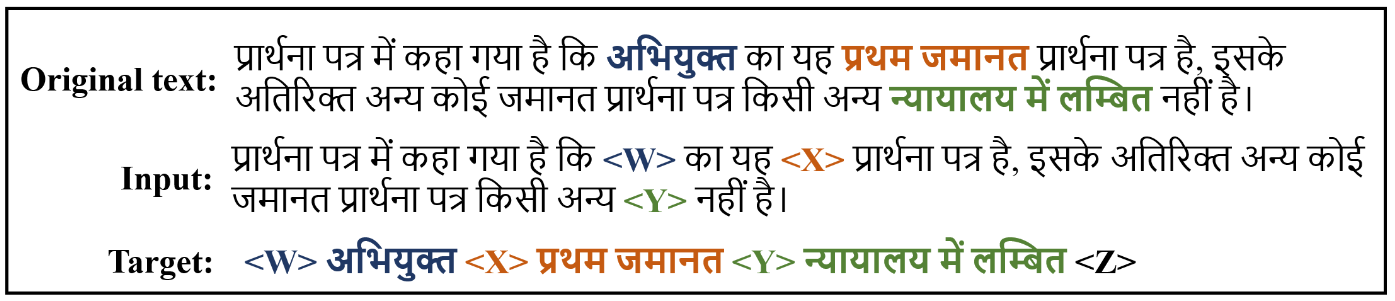}
    \caption{Example of sample data in Hindi for span-corruption denoising pre-training objective.}
\label{fig:pt-hi-example}
\end{figure}

\subsection{Generative Encoder-Decoder Models}
\label{subsec:abs-encodeco-summ}


\noindent \new{As a representative of generative Encoder-Decoder models, we select the \textit{T5}~\citep{T5_paper} model for English-to-English summarization, and its multilingual variant \textit{mT5}~\citep{xue-etal-2021-mt5} for cross-lingual English-to-Hindi legal summarization. 
\textit{T5} is pre-trained over the C4 dataset and treats every NLP task as a text-to-text problem, making it versatile for various applications. 
While \textit{mT5} is pre-trained over the mC4 dataset that covers 101 languages, including Hindi and English, making it well-suited for multilingual and cross-lingual tasks. 
}



\addnew{Figure~\ref{fig:diagram_cpt_sft} demonstrates how we inject legal domain knowledge into generative models and how the different datasets are utilized for this.}
InLegalBERT-PT (for English) and the Bail Corpus (for Hindi) are used for continual pre-training of models. 
Whereas supervised fine-tuning of models is carried out using the train and validation splits of \summcorpus{}. 
The domain-adapted summarization models (specifically for the EN-to-EN and EN-to-HI summarization task over the \summcorpus{} dataset) are obtained through both continual pre-training and supervised fine-tuning of the base models.

\vspace{2mm}
\noindent \textbf{Chunking of long inputs:} The generative models have a limited input capacity (e.g., 512 tokens for \textit{T5}), which presents a challenge since legal judgments are longer in general. To handle this, we employ a chunking strategy by dividing each document into $m$ smaller chunks, where each chunk contains the maximum number of tokens (say, $n$) that the model can process without truncation (e.g., $n$ = 512 for \textit{T5}). Then, the model generates a summary of $t$ tokens for each chunk, and the chunk-wise summaries are concatenated in the same order in which the chunks appear in the document, such that the combined summary (of total $m*t$ tokens) remains aligned with the length of the reference summary.


\vspace{2mm}
\noindent \textbf{Fine-tuning data generation:} As these generative models have limited input capacity, fine-tuning requires document chunks paired with corresponding chunk-specific summaries. 
Let $(d, s)$ represent a (document, summary) pair, where $d$ is segmented into $m$ chunks (i.e., $d_1, d_2, ..., d_m$), where the size of each $d_i$ is the maximum number of tokens that the model can accept without truncating. 
Note that, in the case of cross-lingual summarization, the language of the document ($d$) and summary ($s$) can be different (say, $d$ in English and $s$ in Hindi). 
Now, to generate chunk-specific reference summaries $s_i$ for each $d_i$, we first map every sentence in the summary $s$ to the most similar sentence in $d$ using similarity measure between sentence embeddings (considering the mean of token-level embeddings as the sentence representation). We utilize the LaBSE (Language-agnostic BERT Sentence Embedding)~\citep{LaBSE_paper} model to generate embeddings, given the need to handle both Hindi and English sentences during cross-lingual fine-tuning data generation. For each chunk $d_i$, we then identify all sentences from $s$ that are mapped to any sentence in $d_i$ and use them as the reference summary chunk $s_i$ for that chunk $d_i$. This ensures that the reference summary for each chunk remains contextually relevant. Finally, we get a large number of $(d_i, s_i)$ pairs from each $(d, s)$ pair, resulting in a large dataset for fine-tuning.





\begin{table}[t]
\centering
    \captionsetup{width=\textwidth}
    \caption{Performance of the models for English-to-English summarization. 
    All scores are averaged over the test split of \summcorpus{}. 
    In each class of methods, the scores marked with * indicate that the score is statistically significantly higher than the score of the corresponding baseline (\textit{SummaRuNNer} or FT-only models) for the same metric according to both \textit{Wilcoxon Signed-Rank} and \textit{Mann-Whitney-U} tests at 99\% confidence interval.
    The highest score within each class of methods is highlighted in \textbf{bold}. For the consistency metrics SummaC and NEPrec, the highest score of abstractive summaries has been highlighted.}
    \label{tab:EngSummResults}
    
    \begin{tabular*}{\textwidth}{@{\extracolsep\fill}lccccc@{\extracolsep\fill}}
    \toprule
    \multirow{2}{*} {\textbf{Model}} & \multicolumn{5}{@{}c@{}}{\textbf{English Summary}} \\ \cmidrule{2-6}
    & \textbf{R-2 F1} & \textbf{R-L F1} & \textbf{InL-B-S} & \textbf{SummaC} & \textbf{NEPrec}\\ 
    \hline
    \multicolumn{6}{|l|}{\textbf{Extractive -- Encoder-only}} \\
    \hline
    \textit{SummaRuNNer}$^\dagger$ & 32.27 & 30.34 & 68.91 & 100.0 & 100.0 \\  
    \textit{InLegalSumExt} (our method) & \textbf{38.63}* & \textbf{34.91}* & \textbf{70.50}* & 100.0 & 100.0 \\ 
    \hline
    \multicolumn{6}{|l|}{\textbf{Abstractive -- Encoder-Decoder}} \\
    \hline
    \textit{T5-large-FT} & 30.50 & 27.43 & 69.02 & 67.85 & 65.30 \\ 
    \textit{T5-large-PT-FT} (our method) & \textbf{32.65}* & \textbf{29.13}* & \textbf{69.40} & \textbf{81.80}* & \textbf{83.80}* \\ 
    \hline
    \multicolumn{6}{|l|}{\textbf{Abstractive -- Decoder-only}} \\
    \hline
    \textit{Gemma-2-2b-FT} & 28.20 & 27.38 & 67.20 & 64.81 & 47.00 \\ 
    \textit{Gemma-2-2b-PT-FT} (our method) & \textbf{32.00}* & \textbf{30.12}* & \textbf{70.32}* & \textbf{66.92}* & \textbf{48.32} \\ 
    \hline
    \multicolumn{6}{|l|}{\textbf{Abstractive -- Proprietary LLM}} \\
    \hline
    \textit{GPT-4}$^\ddagger$ & 21.17 & 23.57 & 70.02 & 69.18 & 76.40 \\
    \bottomrule
    \end{tabular*}
    
\vspace{2pt}
{\footnotesize
$^\dagger$ Best score reported in~\citep{datta-etal-2023-mildsum}. \\ 
$^\ddagger$ \textit{GPT-4} is evaluated over 50 random samples from the test split. 
}
\end{table}

\subsubsection{Experiments with \textit{T5} (EN-to-EN)}

\noindent \textbf{Injecting domain knowledge via pre-training and fine-tuning:}
\new{We use the \textit{T5-large}\footnote{\url{https://huggingface.co/google-t5/t5-large}} variant having 770M parameters. 
To enrich its understanding of legal terms and contexts, we first conduct continual pre-training over a randomly selected subset of 27K documents from the InLegalBERT-PT corpus (described in Section~\ref{subsec:pt-data-details}), having around 100K samples, each with 512 tokens. 
We follow the same fill-in-the-blank-style span-corruption denoising task used for original \textit{T5} pre-training~\citep{T5_paper}. 
The schematic representation of this pre-training objective is provided in Figure~\ref{fig:pt-en-example}.
Here, consecutive corrupted tokens form a span, and each span is replaced by a single mask token. 
We mask around 15\% of the tokens randomly from the input sequence of 512 tokens, and the average span length for corruption is set to 3 tokens.\footnote{A larger corruption rate and span length results in longer target sequences, potentially slowing down the training process.}
The model is then trained to predict and recover these missing spans in the input, allowing it to better grasp the intricate structure and nuances of legal texts. We refer to this \textit{T5-large} model that has been continually pre-trained over legal text as \textit{T5-large-PT}.
}

Next, we conduct supervised fine-tuning of both \textit{T5-large} and \textit{T5-large-PT} models using the train and validation splits of \summcorpus{} to tailor the models specifically for the Indian legal summarization task. 
Fine-tuning data is generated by the method described earlier in Section~\ref{subsec:abs-encodeco-summ}. 
We refer to these two fine-tuned models as \textit{T5-large-FT} (the original \textit{T5-large} fine-tuned only) and \textit{T5-large-PT-FT} (\textit{T5-large} pre-trained \& fine-tuned).

\vspace{2mm}
\noindent \textbf{EN-to-EN summarization results:}
\new{The performances of \textit{T5-large-PT-FT} and \textit{T5-large-FT}  over the test split of \summcorpus{} are reported in Table~\ref{tab:EngSummResults}. 
\textit{T5-large-PT-FT} outperforms \textit{T5-large-FT} in terms of both relevance and consistency metrics, demonstrating the superior quality of summaries generated by the \textit{T5-large-PT-FT} model. 
The performance improvement is also statistically significant for both the ROUGE metrics as well as the consistency metrics SummaCONV and NEPrec, according to both \textit{Wilcoxon Signed-Rank} and \textit{Mann-Whitney-U} tests at 99\% confidence interval.
These results demonstrate that domain-specific continual pre-training plays a crucial role in injecting legal domain knowledge and improving model performance. 
}

\begin{table}[t]
\centering
\captionsetup{width=\textwidth}
    \caption{Performances of \textit{T5-large} over the test split of \summcorpus{} after pre-training and fine-tuning. Both GaLore-based and traditional pre-training
    have been done with varying corpus sizes of 27K documents ($\sim$100K samples) and 68K documents ($\sim$250K samples) from the InLegalBERT-PT corpus. All scores are averaged over the test split of \summcorpus{}. The best value of each metric is \textbf{boldfaced}.}
    \label{tab:pt-corpus-results}
    \begin{tabular}{@{}cccccc@{}}
    \toprule
    \multirow{2}{*} {\textbf{Model}} & \textbf{PT} & \textbf{GaLore} & \multicolumn{3}{c}{\textbf{English Summary}} \\ \cmidrule{4-6}
    & \textbf{Size} & \textbf{PT} & \textbf{R-2} & \textbf{R-L} & \textbf{InL-B-S} \\ 
    \midrule
    \multirow{5}{*} {\textit{T5-large-PT-FT}} & \multirow{2}{*} {27K} & No & 32.00 & 28.76 & 69.20 \\ \cmidrule{3-6} 
    & & \textbf{Yes} & \textbf{32.65} & 29.13 & 69.40 \\ 
    \cmidrule{2-6}
    & \multirow{2}{*} {68K} & No & 31.51 & 28.40 & 69.14 \\ \cmidrule{3-6} 
    & & \textbf{Yes} & 32.51 & \textbf{29.24} & \textbf{69.50} \\ 
    \bottomrule
    \end{tabular}
\end{table}

\vspace{2mm}
\noindent \textbf{Performance analysis of pre-training with varying corpus sizes:} To investigate the trade-off between pre-training dataset size and downstream summarization performance, we pre-train on a smaller corpus and a larger corpus.
Specifically, we conduct continual pre-training of T5-large on two randomly selected subsets of the InLegalBERT-PT corpus (described in Section~\ref{subsec:pt-data-details}) -- 
(i)~A subset of 27K documents, having around 100K samples, each with 512 tokens, and 
(ii)~A subset of 68K documents, having around 250K samples, each with 512 tokens. 
We finetune both pre-trained models for the downstream legal summarization task and compare their performances. 
Our findings, shown in Table~\ref{tab:pt-corpus-results}, show that the performance achieved with the smaller subset is nearly identical to that with the larger subset.
Hence, we choose the smaller subset (having 27K documents) for experiments with other abstractive models, ensuring a more efficient pre-training process without compromising performance much.

\vspace{2mm}
\noindent \textbf{Exploration of memory-efficient pre-training strategies:}
In pre-training, we also explored both traditional full-parameter pre-training and the GaLore (Gradient Low-Rank Projection) training strategy~\citep{galore_paper}, which enables memory-efficient full-parameter learning, making it a promising alternative to traditional pre-training. 
The results, presented in Table~\ref{tab:pt-corpus-results}, demonstrate that GaLore-based pre-training shows slightly better performance in the downstream summarization task.

Given these observations, we chose the \textit{T5-large} model pre-trained over 27K legal documents using GaLore training for most experiments in this paper. This model, referred to as \textit{T5-large-PT}, is the one reported in Table~\ref{tab:EngSummResults}.

\subsubsection{Experiments with \textit{mT5} (EN-to-HI)}

We select the \textit{mT5-large}\footnote{\url{https://huggingface.co/google/mt5-large}} model having 1B parameters for cross-lingual summarization of English legal documents into Hindi. 
Here, we explore comprehensive multilingual pre-training experiments to investigate cross-lingual transfer effects and the impact of domain-specific multilingual pre-training.

\vspace{2mm}
\noindent \textbf{Injecting domain knowledge via multilingual pre-training and fine-tuning:} We perform continual pre-training in 3 settings, to understand how different approaches compare: 
\begin{enumerate}[leftmargin=*, label=(\roman*), nolistsep]
    \item \textit{Pre-training (PT) over English(EN)-only legal texts:} $\sim$100K samples (each with 512 tokens) generated from 27K English documents of the InLegalBERT-PT corpus. 
    \item \textit{PT over Hindi(HI)-only legal texts:} $\sim$100K samples (each with 512 tokens) generated from 34K Hindi documents of the Bail corpus.
    \item \textit{PT over EN+HI (equally mixed) legal texts:} $\sim$50K samples (each with 512 tokens) generated from 14K English documents of the InLegalBERT-PT corpus and $\sim$50K samples (each with 512 tokens) generated from 17K Hindi documents of the Bail corpus.
\end{enumerate}


\noindent In all 3 settings, the number of pretraining samples is the same, i.e., $\sim$100K samples (each with 512 tokens), and we use the same span-corruption denoising objective for the continual pre-training (illustrated in Figures~\ref{fig:pt-en-example} and \ref{fig:pt-hi-example}).
Around 15\% of the tokens are randomly masked from an input sequence of 512 tokens, with an average corruption span length of 3 tokens. 
Since GaLore pre-training performs better for \textit{T5}, we have applied the same memory-efficient GaLore training strategy here. 
Finally, after continual pre-training, the domain-adapted pre-trained checkpoints from these three settings are referred to as \textit{mT5-large-PT (EN-only PT)}, \textit{mT5-large-PT (HI-only PT)}, and \textit{mT5-large-PT (EN+HI PT)}.

We fine-tune all three pre-trained models as well as the \textit{mT5-large} (without pre-training) for the downstream cross-lingual English-to-Hindi summarization task.
Fine-tuning data is generated from the cross-lingual (English document - Hindi summary pairs) train split of \summcorpus{} to further align them with the cross-lingual summarization of Indian legal documents.
These fine-tuned models are referred to as \textit{mT5-large-PT-FT (EN-only PT)}, \textit{mT5-large-PT-FT (HI-only PT)}, \textit{mT5-large-PT-FT (EN+HI PT)}, and \textit{mT5-large-FT} (no domain-specific pre-training).

\begin{table}[t]
\centering
    \caption{Performances for English-to-Hindi summarization using generative models.
    All scores are averaged over the test split of \summcorpus{}.
    The scores with * indicate a score that is statistically significantly higher than the score of the \textit{mT5-large-FT} for the same metric in terms of both \textit{Wilcoxon Signed-Rank} and \textit{Mann-Whitney-U} tests at a 99\% confidence interval. Best value of each metric is \textbf{boldfaced}.    
    }
    \label{tab:HinSummResults}
    
    \begin{tabular}{lccc}
    \toprule
    \multirow{2}{*} {\textbf{Model}} & \multicolumn{3}{c}{\textbf{Hindi Summary}} \\ \cmidrule{2-4}
    & \textbf{R-2} & \textbf{R-L} & \textbf{B-S} \\ 
    \midrule
    \textit{CrossSum-mT5-FT}$^\dagger$ & 21.76 & 20.68 & 75.05 \\ \hline \hline
    \textit{mT5-large-FT} & 22.84 & 21.82 & 74.15 \\ \hline
    \textit{mT5-large-PT-FT (EN-only PT)} & 24.98* & 23.03* & 74.40 \\ \cmidrule{2-4}
     \textit{mT5-large-PT-FT (HI-only PT)} & 26.40* & 24.39* & 75.01* \\ \cmidrule{2-4}
     \textit{mT5-large-PT-FT (EN+HI PT)} & \textbf{26.68}* & \textbf{24.61}* & \textbf{75.12}* \\ 
    \hline 
    \hline
    \textit{GPT-4}$^\ddagger$ & 07.10 & 13.26 & 69.07 \\ 
    \bottomrule
    \end{tabular}
    
\vspace{3pt}
{\footnotesize
$^\dagger$ Best score reported in~\citep{datta-etal-2023-mildsum}. \\ 
$^\ddagger$ \textit{GPT-4} is evaluated over 50 random samples from the test split. 
}
\end{table}

\vspace{2mm}
\noindent \textbf{EN-to-HI summarization results:}
The performances of these models over cross-lingual test split of \summcorpus{} are presented in Table~\ref{tab:HinSummResults}.
Several key observations arise from this analysis: 
\begin{enumerate}[leftmargin=*, label=(\roman*), nolistsep]
    \item \textit{EN-only Pretraining (Setting-i)}: Continual pretraining on English-only legal texts [\textit{mT5-large-PT-FT (EN-only PT)}] also improves the English-to-Hindi legal summarization, compared to fine-tuning alone (\textit{mT5-large-FT}), demonstrating the utility of domain(legal)-specific knowledge even in a single language. 
    \item \textit{HI-only Pretraining (Setting-ii)}: Pretraining on Hindi-only legal texts [\textit{mT5-large-PT-FT (HI-only PT)}] further enhances performance, showing the benefits of language-specific domain adaptation.
    \item \textit{EN+HI Pretraining (Setting-iii)}: Finally, pre-training over English+Hindi (equally mixed) legal texts [\textit{mT5-large-PT-FT (EN+HI PT)}] achieves the best results, leveraging knowledge from both source (English) and target (Hindi) languages for the English-to-Hindi summarization task.
\end{enumerate}

\noindent These results confirm the positive cross-lingual transfer effects of domain-specific multilingual pre-training, highlighting the importance of utilizing both source (English) and target (Hindi) language legal corpora for cross-lingual summarization tasks. 

Our best-performing model, \textit{mT5-large-PT-FT (EN+HI PT)} demonstrates 23\% improvement in ROUGE-2 F1 and 19\% improvement in ROUGE-L F1 scores compared to \textit{CrossSum-mT5-FT}, which was the SOTA cross-lingual summarization results on MILDSum~\citep{datta-etal-2023-mildsum}. 
Notably, both \textit{mT5-large-PT-FT} and \textit{CrossSum-mT5-FT} are fine-tuned over \summcorpus{}. The original \textit{CrossSum-mT5}\footnote{\url{https://huggingface.co/csebuetnlp/mT5_m2o_hindi_crossSum}} was created by fine-tuning \textit{mT5} on a large collection of cross-lingual document-summary pairs where the target summary is in Hindi~\citep{bhattacharjee2023crosssum}.
These results highlight that \textit{task-agnostic pre-training on a small amount of domain(legal)-specific unlabeled data leads to better downstream summarization performance than training the model with a large general-domain task(summarization)-specific labeled data} as was done in \citet{bhattacharjee2023crosssum}.
\subsection{Generative Decoder-only Models}
\label{subsec:abs-deco-summ}

As a representative of recent open-source generative models, we select \textit{Gemma-2}~\citep{Riviere2024Gemma2I} for our experiments on English-to-English legal document summarization. 
Specifically, to account for resource constraints, we choose the 2B parameter variant -- \textit{Gemma-2-2b}\footnote{\url{https://huggingface.co/google/gemma-2-2b}}.


\vspace{2mm}
\noindent \textbf{Experiments with \textit{Gemma-2}:}
To enhance the understanding of the Indian Law domain, we first conduct continual pre-training of \textit{Gemma-2-2b} on the same 27K English documents of InLegalBERT-PT corpus.
Unlike the span-corruption objective used in \textit{T5}, here we follow the \textit{Next Token Prediction} pre-training objective.
Similar to our pre-training experiments conducted with T5, we explored both settings here: pre-training all model parameters, without and with GaLore. 
In both cases, we consider the input sequences of 128 tokens (resulting in $\sim$400K samples) to ensure efficient learning with a higher batch size of 32. 

\begin{table}[t]
\centering
    \captionsetup{width=\textwidth}
    \caption{Comparison of summarization results between GaLore pre-training vs. traditional pre-training for the Decoder-only model (\textit{Gemma-2}) over \summcorpus{}. All scores are averaged over the test split of \summcorpus{}. The highest score for each metric is highlighted in \textbf{bold}.}
    \label{tab:galore-results}
    \begin{tabular}{@{}lcccc@{}}
    \toprule
    \multirow{2}{*} {\textbf{Model}} & \textbf{GaLore} & \multicolumn{3}{c}{\textbf{English Summary}} \\ \cmidrule{3-5}
    & \textbf{PT} & \textbf{R-2} & \textbf{R-L} & \textbf{InL-B-S} \\ 
    \hline
    \multirow{2}{*} {\textit{Gemma-2-2b-PT-FT}} & No & 31.38 & 29.74 & 69.80 \\ \cmidrule{2-5} 
    & \textbf{Yes} & \textbf{32.00} & \textbf{30.12} & \textbf{70.32} \\ 
    \bottomrule
    \end{tabular}
\end{table}

\vspace{2mm}
\noindent \textbf{Effectiveness of GaLore pre-training:}
In line with our experiments on the Encoder-Decoder architecture, we assess the effectiveness of the GaLore pre-training compared to traditional full-parameter pre-training for this Decoder-only model, \textit{Gemma-2-2b}.
Our findings in Table~\ref{tab:galore-results} show that for \textit{Gemma-2-2b} also, GaLore-based pre-training performs slightly better than the conventional pre-training on the downstream summarization task.

Given the larger context window of \textit{Gemma-2}, we conduct supervised fine-tuning of both \textit{Gemma-2-2b-PT} (domain-adapted) and \textit{Gemma-2-2b} models with a maximum sequence length of 5,632 tokens ($\sim$4K document chunk + $\sim$1.5K summary chunk), enabling it to better handle the lengthy nature of legal documents. To this end, using the method described in Section~\ref{subsec:abs-encodeco-summ}, fine-tuning data is generated by chunking documents into segments of 4K tokens. 
\new{Due to the large size of 2B parameters, we prioritized efficient training methods to optimize both computational costs and performance. We employed a parameter-efficient tuning technique, LoRA (Low-Rank Adaptation)~\citep{Hu2021LoRALA}, for parameter-efficient fine-tuning.
}

\vspace{2mm}
\noindent \textbf{Summarization results:} \new{The performances of \textit{Gemma-2-2b-PT-FT} (pre-trained and fine-tuned) and \textit{Gemma-2-2b-FT} (fine-tuned only) over the test split of \summcorpus{} are reported in Table~\ref{tab:EngSummResults}. \textit{Gemma-2-2b-PT-FT} performs statistically significantly better than \textit{Gemma-2-2b-FT} for most of the relevance and consistency metrics. These results further highlight the importance of domain-specific continual pre-training.
}

\subsection{Details of the hyperparameters configuration}
\label{subsec:hyper-param}

The hyperparameters used for pre-training and fine-tuning experiments and their specific configurations for different generative models are reported in this section. Pre-training hyperparameters for \textit{T5-large},
\textit{mT5-large},
and \textit{Gemma-2-2b},
are tabulated in Table~\ref{tab:pt-param-appendix}, where hyperparameters for GaLore pre-training are also reported. Table~\ref{tab:ft-param-appendix} shows the hyperparameters for fine-tuning along with the LoRA hyperparameters used for \textit{Gemma-2-2b}. The GPU time consumed for each of the continual pre-training and supervised fine-tuning experiments for \textit{T5-large}, \textit{mT5-large}, and \textit{Gemma-2-2b} are detailed in the Appendix~\ref{subsec:gpu-time-appendix}.

\begin{table}[t]
    \caption{Hyperparamters used in pre-training of \textit{T5-large}, \textit{mT5-large}, and \textit{Gemma-2-2b}.}
    \label{tab:pt-param-appendix}
    \begin{tabular*}{\textwidth}{@{\extracolsep\fill}lcl}
    \toprule
    \textbf{Model} & \textbf{GaLore PT} & \textbf{Hyperparameters for Pre-training} \\
    \toprule
    \multirow{5}{*} {\textit{T5-large}} & \multirow{2}{*} {No} & Batch: 128, Learning rate: 5e-4, Epoch: 3, \\
    & & Max input length: 512, Max output length: 128 \\
    \cmidrule{2-3}
    & \multirow{3}{*} {Yes} & Batch: 128, Learning rate: 5e-4, Epoch: 3, \\
    & & Max input length: 512, Max output length: 128 \\
    & & GaLore:\{r: 512, update\_proj\_gap: 500, scale: 0.5\}\\
    \toprule
    \multirow{3}{*} {\textit{mT5-large}} & \multirow{3}{*} {Yes} & Batch: 32, Learning rate: 1e-4, Epoch: 3, \\
    & & Max input length: 512, Max output length: 192 \\
    & & GaLore:\{r: 512, update\_proj\_gap: 500, scale: 0.5\}\\
    \toprule
    \multirow{5}{*} {\textit{Gemma-2-2b}} & \multirow{2}{*} {No} & Batch: 16, Learning rate: 1e-5, Epoch: 2, \\
    & & Max Seq. length: 128 \\
    \cmidrule{2-3}
    & \multirow{3}{*} {Yes} & Batch: 32, Learning rate: 5e-5, Epoch: 2, \\
    & & Max Seq. length: 128, GaLore:\{r: 512, \\
    & & update\_proj\_gap: 500, scale: 0.5\} \\
    \bottomrule
    \end{tabular*}
\end{table}

\subsection{Comparison with \textit{GPT-4}}
\label{subsec:gpt-4}

We check the performance of \textit{GPT-4}, arguably the SOTA LLM, for both monolingual (EN-to-EN) and cross-lingual (EN-to-HI) summarization of Indian legal documents. 
Given the relatively high subscription costs of GPT-4, we evaluated it over 50 randomly selected documents from the test split of \summcorpus{}.

\vspace{2mm}
\noindent \textbf{Prompts for \textit{GPT-4}:} The following instruction-based prompt is used with \textit{GPT-4} for generating English summaries from English legal documents (EN-to-EN summarization): 
``\textit{Please generate a summary of length at most $<$max\_length$>$ words for the following Indian Legal judgment: [\textbackslash n]$<$Judgment\_in\_English$>$ [\textbackslash n \textbackslash n] Summary:}'' (where the parameter  \textit{$<$max\_length$>$} is decided from the reference summary of the particular document). 

For cross-lingual EN-to-HI summarization, we have used the following instruction-based prompt to generate the Hindi summaries in the Devanagari script from English case judgments:
``\textit{Please generate a Hindi summary in Devanagari script of length at most $<$max\_length$>$ words for the following Indian Legal judgment: [\textbackslash n]$<$Judgment\_in\_English$>$ [\textbackslash n \textbackslash n] Summary in Hindi in Devanagari script:}'' (where the parameter \textit{$<$max\_length$>$} is decided from the reference summary of the particular document).

\begin{table}[t]
    \centering
    \captionsetup{width=\textwidth}
    \caption{Hyperparamters used in fine-tuning of \textit{T5-large}, \textit{mT5-large}, and \textit{Gemma-2-2b}.}
    \label{tab:ft-param-appendix}
    \begin{tabular}{@{}ll@{}}
    \toprule
    \textbf{Model} & \textbf{Hyperparameters for Fine-tuning} \\
    \toprule
    \multirow{3}{*} {\textit{T5-large}} & Batch: 8, Learning rate: 5e-5, Max Epoch: 10, \\
    & EarlyStop Patience: 2, Max input length: 512, \\
    & Max output length: 192 \\
    \toprule
    \multirow{3}{*} {\textit{mT5-large}} & Batch: 8, Learning rate: 5e-5, Max Epoch: 10, \\
    & EarlyStop Patience: 2, Max input length: 512, \\
    & Max output length: 256 \\
    \toprule
    \multirow{4}{*} {\textit{Gemma-2-2b}} & Batch: 1, Learning rate: 1e-5, Max Epoch: 5, \\
    & EarlyStop Patience: 2, Max Seq. length: 5632, \\
    & Q-LoRA:\{load\_in\_4bit: \textit{True}, r: 8, \\
    & $\alpha$: 32, dropout: 0.05\} \\
    \bottomrule
    \end{tabular}
\end{table}

\vspace{2mm}
\noindent \textbf{EN-to-EN summarization results:} 
The performance of \textit{GPT-4} is reported in Table~\ref{tab:EngSummResults}. 
The metric values for \textit{GPT-4} are notably lower than the domain(legal)-specific \textit{FT} and \textit{PT-FT} models in terms of ROUGE (R-2 \& R-L) and factual consistency metrics (SummaC \& NEPrec). 

\vspace{2mm}
\noindent \textbf{EN-to-HI summarization results:} 
As observed in Table~\ref{tab:HinSummResults}, the cross-lingual performance of \textit{GPT-4} lags significantly behind the domain(legal)-specific \textit{FT} and \textit{PT-FT} models. In particular, our best-performing model, \textit{mT5-large-PT-FT (EN+HI PT)} demonstrates more than 19 points improvement in R-2, 11 points improvement in R-L, and 6 points improvements in B-S over \textit{GPT-4}. 

\noindent \new{Hence, we see that \textit{GPT-4} still struggles with domain-specific and cross-lingual tasks. In contrast, models tailored to the legal domain through domain-adaptive continual pre-training plus supervised fine-tuning exhibit superior performance, validating the importance of domain-specific adaptation. 
}

\section{Expert evaluation of summaries}
\label{sec:expert-eval}

To further validate the quality of the model-generated summaries, we conducted an expert evaluation via a Law domain expert. 
The expert is an LLB graduate, with two years of professional experience, and currently pursuing LLM at one of the most reputed Law schools in India (details omitted for anonymity). 
The expert was selected based on the recommendation of a Professor from the said law school, ensuring two key criteria: (i)~a strong understanding of the Indian legal domain, and (ii)~proficiency in both English and native fluency in Hindi. 
The expert was informed of the purpose of the survey and was paid a mutually agreed honorarium for the evaluation. 


\subsection{Survey setup for Expert Evaluation}

Given the limited availability of the Law domain expert, we conducted this evaluation on 10 randomly selected judgments from the test split of \summcorpus{} dataset. 
Specifically, evaluation was performed over 40 (Judgment-J, Summary-S) pairs, comprising 4 summaries for each of the 10 judgments -- 2 summaries in English generated by \textit{Gemma-2-2b-FT} and \textit{Gemma-2-2b-PT-FT}, and 2 summaries in Hindi generated by \textit{mT5-large-FT} and \textit{mT5-large-PT-FT (EN+HI PT)}). 
To eliminate potential bias in the evaluation, we did \textit{not} disclose to the expert which summary is generated by which model.

The expert was asked to evaluate a $(J, S)$ pair on a Likert scale of 1-5 (with 1 being the lowest / worst score and 5 being the highest / best score) for each of the following metrics: 

\noindent (i)~\underline{\textit{Preservation of Important Information}} (IMP): how well the most important information in $J$ has been covered in $S$, 

\noindent (ii)~\underline{\textit{Factual Consistency Score}} (FCS): how well the facts and information in $S$ are consistent with those in $J$, 

\noindent (iii)~\underline{\textit{Fluency}} (FLY): grammatical accuracy, coherence, and smooth flow of $S$, and 

\noindent (iv--viii)~\underline{\textit{Coverage of Rhetorical Roles}}: how well the summary $S$ represents five key rhetorical roles / segments generally present in an Indian judgment, i.e., \textit{Facts} (FAC), \textit{Argument} (ARG), \textit{Statute and Precedent citations} (CIT), \textit{Ratio of the Decision} (RTO), and \textit{Ruling by Present Court} (RPC). 
The expert was asked to rate the summary for each rhetorical role on a Likert scale of 1-5, where 1 means the role is minimally captured in the summary, and 5 means the role is accurately captured and well-represented in the summary.

For more details about these rhetorical roles, please refer to \citep{rhetorical_role_paper}. Details about the full instructions given to the expert, are given in Appendix~\ref{sec:expert-eval-appendix}.


\subsection{Results of Expert Evaluation}

Expert evaluation results are presented in Table~\ref{tab:expert-results}. For English-to-English summarization, the expert consistently preferred the summaries generated by \textit{Gemma-2-2b-PT-FT} to its \textit{FT}-only counterpart across most metrics, particularly in IMP, FCS, and in four out of five rhetorical roles. 
For English-to-Hindi summarization also, the summaries by the \textit{PT+FT} model have been preferred over the \textit{FT}-only model across FCS, FLY, and three out of the five rhetorical roles. 
Overall, the findings from the expert evaluation emphasize the effectiveness of the legal domain-specific continual pretraining of the \textit{PT+FT} models.

\begin{table}[t]
\centering
    \captionsetup{width=\textwidth}
    \caption{Expert evaluation scores for English-to-English and English-to-Hindi summaries generated by \textit{PT+FT} and \textit{FT}-only models, averaged over 10 randomly selected judgments from the test split of \summcorpus{}. The scores for each metric range from [1-5], with 1 being the worst / lowest and 5 being the best / highest.
    }
    \label{tab:expert-results}
    \begin{tabular*}{\textwidth}{@{\extracolsep\fill}lccc||ccccc@{\extracolsep\fill}}
    \hline
    \textbf{Model} & \textbf{IMP} & \textbf{FCS} & \textbf{FLY} & \textbf{FAC} & \textbf{ARG} & \textbf{CIT} & \textbf{RTO} & \textbf{RPC} \\
    \hline
    \multicolumn{9}{|l|}{\textbf{English-to-English Summarization}} \\
    \hline
    \textit{Gemma-2-2b-FT} & 3.0 & 3.5 & 5.0 & 3.5 & 2.0 & 2.5 & 3.3 & 4.5 \\ 
    \midrule
    \textit{Gemma-2-2b-PT-FT} & \textbf{4.5} & \textbf{4.0} & 5.0 & \textbf{4.2} & \textbf{3.5} & \textbf{3.5} & \textbf{4.2} & 4.5 \\
    \hline
    \multicolumn{9}{|l|}{\textbf{English-to-Hindi Summarization}} \\
    \hline
    \textit{mT5-large-FT} & 4.0 & 3.8 & 2.5 & 2.8 & \textbf{3.8} & 3.3 & 3.5 & 3.5 \\ 
    \midrule
    \textit{mT5-large-PT-FT (EN+HI PT)} & 4.0 & \textbf{4.2} & \textbf{3.7} & \textbf{4.0} & 3.5 & 3.3 & \textbf{3.7} & \textbf{4.0} \\ 
    \hline
    \end{tabular*}
\end{table}

\vspace{2mm}
\noindent \textbf{Error analysis:} If the expert gave low scores (3 or less) for any of the metrics described above, then the expert was asked to give a justification for the lower score. Based on the expert's comments, we make the following observations:
(i)~For Hindi summaries generated by \textit{mT5-large-FT}, the sentences lacked coherence, affecting the FLY score, though legal terminologies were used accurately. In comparison, the corresponding summaries of \textit{mT5-large-PT-FT (EN+HI PT)} were observed to be more fluent.
(ii) The expert observed some \textit{hallucinations} in \textit{mT5-large-FT}'s Hindi summaries, e.g., the model incorrectly mentioned `\textit{Supreme Court}' which is not referenced in the original judgment; in contrast, the \textit{PT+FT} model's summary accurately represented most of the facts, aligning well with the judgment text. 
(iii) For English summaries, both \textit{FT}-only and \textit{PT+FT} models produced fluent summaries and successfully captured the final ruling of the present court. However, the summary generated by \textit{Gemma-2-2b-FT} sometimes misses important facts and statutes, affecting the FAC and CIT scores. In contrast, \textit{Gemma-2-2b-PT-FT}'s summaries demonstrate better coverage of facts and relevant statutes.

\begin{figure}[!htb]
\centering
\captionsetup{width=\textwidth}
\includegraphics[width=\textwidth]{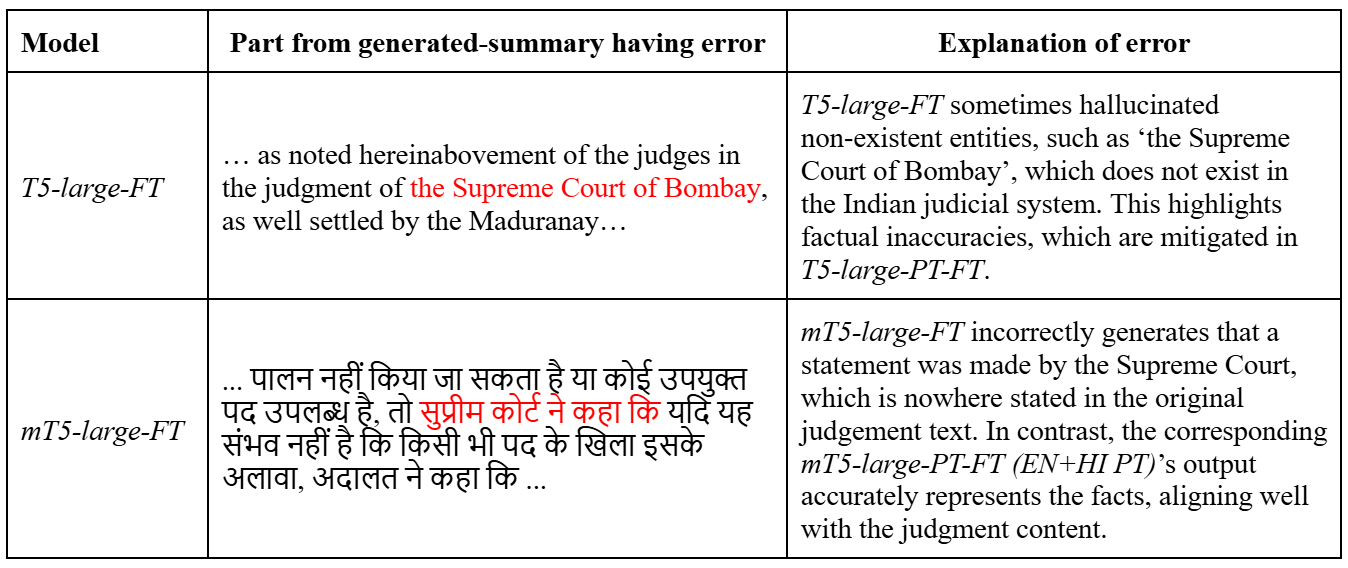}
\caption{Examples of errors in English and Hindi summaries, committed by the \textit{FT}-only models, which are mitigated in the corresponding \textit{PT+FT} models. The last column explains the error.}
\label{fig:error_ft}
\end{figure}


Figure~\ref{fig:error_ft} illustrates some examples of errors observed in both English and Hindi summaries, committed by \textit{FT}-only models, which are mitigated in the corresponding \textit{PT+FT} models. 

\section{Conclusion}
\label{sec:conclu}



\new{
Through our experiments, we demonstrate that injecting legal domain knowledge into generative Encoder-Decoder and Decoder-only models through continual pre-training on legal corpora, enhances their ability to summarize legal documents in both monolingual (English-to-English) and cross-lingual (English-to-Hindi) settings. 
Our results demonstrate that our approaches improve the models' understanding of complex legal texts, leading to more contextually accurate summaries.
We also find that domain-specific extractive summarization benefits from our proposed framework of incorporating domain-specific pre-trained Encoders. 
Our comprehensive evaluation of diverse model architectures highlights the effectiveness of domain knowledge injection across different paradigms.
Our analysis of pre-training strategies reveals that memory-efficient training with comparatively smaller pre-training corpora can achieve nearly equivalent downstream summarization performance comparable to resource-intensive setups. 
Additionally, our investigation into cross-lingual transfer effects finds that models pre-trained on both source and target language corpora perform better in cross-lingual summarization tasks, highlighting the benefits of multilingual exposure.
}

\new{Our models significantly outperform state-of-the-art baselines on the \summcorpus{} benchmark for both English-to-English and English-to-Hindi summarization of Indian legal documents. These improvements are statistically significant in both standard relevance metrics and consistency metrics. 
Moreover, human evaluation by a law domain expert further validates the quality and factual consistency of the summaries generated by our models.
}

\new{
Overall, we believe this study contributes to making legal information more accessible and comprehensible to a broader audience in India, ultimately enhancing access to justice among Indians.
Furthermore, though this paper focuses on summarizing Indian court judgments in English and Hindi, the methods we explore for injecting domain knowledge would be useful for improving legal document summarization in any other language as well.
}


\bibliography{references}
\bibliographystyle{iclr2026_conference}


\clearpage

\noindent {\bf \large Appendix}

\appendix

\section{More details about the experiments}
\label{sec:exp-appendix}

\subsection{GPU time consumed for experiments}
\label{subsec:gpu-time-appendix}
All continual pre-training and supervised fine-tuning experiments were conducted on a system having a NVIDIA RTX A6000 48GB GPU. The GPU time required for continual pre-training was approximately 18 hours for \textit{T5-large}, 20 hours for \textit{mT5-large}, and 18 hours for \textit{Gemma-2.} For supervised fine-tuning over the \summcorpus{} dataset, \textit{T5-large} took around 8 hours, \textit{mT5-large} took around 12 hours and \textit{Gemma-2} required approximately 12 hours. During inference on the test split of \summcorpus{} dataset, the GPU time was approximately 7 hours for \textit{T5-large}, 10 hours for \textit{mT5-large}, and 15 hours for \textit{Gemma-2}.

\section{More details about Expert evaluation}

\subsection{Full instruction given to the Law expert for evaluating summaries}
\label{sec:expert-eval-appendix}

In this evaluation, you will be given a Court Judgment (full text of the Court judgment) in English and four AI-generated summaries generated by four different AI models, two in English and two in Hindi. 

You need to judge the quality of each given summary and give a score for the first three metrics (IMP, FCS, FLY) on a Likert scale of 1--5, with 1 being the lowest/poorest score and 5 being the highest/best score. 

I. \textit{Preservation of Important Information (IMP)}: how well the most critical and important information in the case judgment has been covered in the summary. [Guiding questions: Does the summary include the essential information necessary to understand the judgment?]

II. \textit{Factual Consistency Score (FCS)}: how well the facts, information, and decisions in the summary are consistent with those in the original judgment. Here, you need to focus on identifying factual inaccuracies, contradictions, and the presence of hallucinated (i.e., something that does NOT exist in the original judgment) information in the summary. [Guiding questions: Are the facts, information, and decisions mentioned in the summary consistent with those in the original judgment? Does the summary contain any factual inaccuracies or hallucinated information?]

III. \textit{Fluency (FLY)}: Measures grammatical accuracy, coherence, and smooth flow of the summary text. [Guiding questions: Is the summary free of grammatical and spelling errors (mainly in the named entities)? Does the text flow naturally and cohesively?]

IV. \textit{Coverage of Rhetorical Roles}: Rate the summary based on how well it covers some key Rhetorical roles/segments that are typically present in an Indian court judgment. You have to give a score for each role on a Likert scale of 1-5, where 1 means the role is minimally captured in the summary, and 5 means the role is accurately captured and well-represented in the summary. 
Additionally, if a certain rhetorical segment is completely missing in the summary, give a score of $0$ for that rhetorical role.
The rhetorical roles are described below:

\begin{enumerate}[leftmargin=*, nolistsep]

\item \textit{Facts} (abbreviated as FAC): Refers to the chronology of events that led to filing the case.
\item \textit{Argument} (ARG): Arguments of the contending parties.
\item \textit{Statute and Precedent citations} (CIT): Established Laws, Acts, and prior cases that were referred to by the present judgement.
\item \textit{Ratio of the decision} (RTO): Reasoning / rationale for the final judgment given by the present court.
\item \textit{Ruling by Present Court} (RPC): the final decision / conclusion of the present court.

\end{enumerate}

For any of the above metrics, if you give low scores (3 or less), please mention the reasons in the `Comments' section, e.g., by indicating the errors in the summary.

\end{document}